\documentclass[conference]{IEEEtran}
\IEEEoverridecommandlockouts

\usepackage{cite}
\usepackage{amsmath,amssymb,amsfonts}
\usepackage{algorithmic}
\usepackage{graphicx}
\usepackage{textcomp}
\usepackage{xcolor}
\usepackage{booktabs}   
\usepackage{multirow}   
\usepackage{amssymb}    
\usepackage{booktabs}   
\usepackage{makecell}   
\usepackage{colortbl}   
\usepackage{xcolor}     
\usepackage{tabularx}
\usepackage{soul}       
\usepackage{xcolor}     

\usepackage{fancyhdr} 
\usepackage{lipsum} 

\definecolor{colorLeft}{HTML}{A7C7E7}   
\definecolor{colorRight}{HTML}{FCD5B4}  
\definecolor{colorMiddle}{HTML}{C1E1C1} 
\definecolor{colorUpper}{HTML}{F8B8D4} 
\definecolor{colorLower}{HTML}{FDFFB6} 
\usepackage{graphicx}
\usepackage{booktabs}
\usepackage{tabularx}   
\usepackage{soul}       
\usepackage{xcolor}     
\def\BibTeX{{\rm B\kern-.05em{\sc i\kern-.025em b}\kern-.08em
    T\kern-.1667em\lower.7ex\hbox{E}\kern-.125emX}}
\fancypagestyle{firstpage}{ 
  \fancyhf{} 
  \fancyhead[C]{2025 International Conference on Bioinformatics and Biomedicine (BIBM)} 
}
    
\begin{document}

\title{Spatial-aware Symmetric Alignment for Text-guided Medical Image Segmentation\\
}

\author{\IEEEauthorblockN{Linglin Liao}
\IEEEauthorblockA{\textit{School of Information Engineering, } \\
\textit{Capital Normal University}\\
Beijing, China \\
2231002051@cnu.edu.cn}
\and
\IEEEauthorblockN{Yu Liu}
\IEEEauthorblockA{\textit{School of Information Engineering, } \\
\textit{Capital Normal University}\\
Beijing, China \\
liuyu@cnu.edu.cn}
\and
\IEEEauthorblockN{Qichuan Geng\textsuperscript{*}\thanks{* Corresponding author.}}
\IEEEauthorblockA{\textit{School of Information Engineering, } \\
\textit{Capital Normal University}\\
Beijing, China \\
gengqichuan1989@cnu.edu.cn}
}


\maketitle

\thispagestyle{firstpage} 

\begin{abstract}

Text-guided Medical Image Segmentation has shown considerable promise for medical image segmentation, with rich clinical text serving as an effective supplement for scarce data.
However, current methods have two key bottlenecks. On one hand, they struggle to process diagnostic and descriptive texts simultaneously, making it difficult to identify lesions and establish associations with image regions. On the other hand, existing approaches focus on lesions description and fail to capture positional constraints, leading to critical deviations. Specifically, with the text “in the left lower lung”, the segmentation results may incorrectly cover both sides of the lung.
To address the limitations, we propose the Spatial-aware Symmetric Alignment (SSA) framework to enhance the capacity of referring hybrid medical texts consisting of locational, descriptive, and diagnostic information. Specifically, we propose symmetric optimal transport alignment mechanism to strengthen the associations between image regions and multiple relevant expressions, which establishes bi-directional fine-grained multimodal correspondences. In addition, we devise a composite directional guidance strategy that explicitly introduces spatial constraints in the text by constructing region-level guidance masks. Extensive experiments on public benchmarks demonstrate that SSA achieves state-of-the-art (SOTA) performance, particularly in accurately segmenting lesions characterized by spatial relational constraints.

\end{abstract}

\section{Introduction}
\label{sec:introduction}
Text-guided medical image segmentation aims to delineate target regions in medical images according to textual descriptions~\cite{chen2024causalclipseg,li2023lvit}. Despite its progress, this paradigm still suffers from a significant semantic gap between visual and textual modalities, making it difficult to achieve precise cross-modal alignment for fine-grained lesion localization.

\begin{figure}[t]
    \centering
    \includegraphics[width=1\columnwidth]{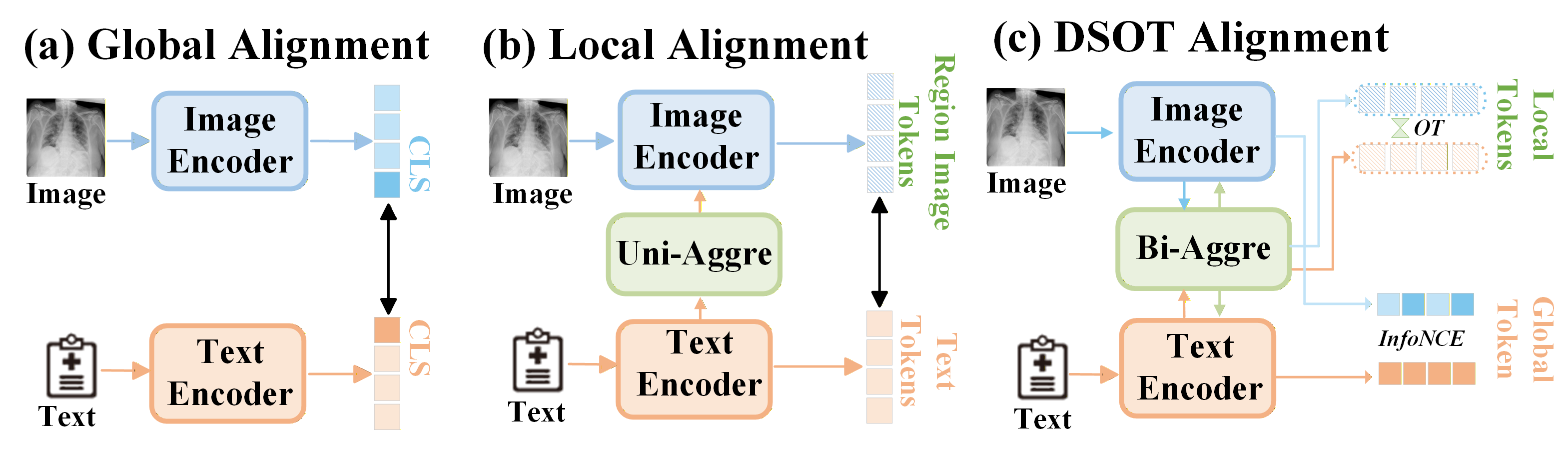}
    \caption{Existing multi-modal methods perform alignment by adopting either (a) Global Alignment or (b) Local Alignment based on uni-aggregation. In this paper, we propose (c) DSOT alignment, which performs both global and local alignment on bi-aggregated features, implemented via InfoNCE and symmetric OT algorithms respectively.}
    \label{fig:my_image_label}
\end{figure}

Global and local alignment paradigms have been widely explored to bridge the semantic gap of multimodal data. Global alignment methods~\cite{radford2021learning} establish coarse correspondences between the entire image and the text, as illustrated in Fig.~\ref{fig:my_image_label}(a). Local alignment methods~\cite{huang2021gloria,cha2023learning} focus on region-level matching for finer associations, as illustrated in Fig.~\ref{fig:my_image_label}(b). Despite these advances, current methods still face two bottlenecks. Existing methods struggle to process medical texts that simultaneously contain diagnostic and descriptive information, making it difficult to accurately identify lesion and to establish consistent associations with image regions. At the same time, they primarily focus on lesion-centric descriptions while failing to capture positional constraints, which leads to critical deviations. 

To overcome these limitations, we propose the \textbf{Spatial-aware Symmetric Alignment (SSA) framework}, which enhances the capacity of referring hybrid medical texts consisting of locational, descriptive, and diagnostic information. To strengthen the associations between lesion regions and various textual referring, we design a \textbf{Dual-granularity Symmetric Optimal Transport (DSOT) alignment algorithm}. As illustrated in Fig.~\ref{fig:my_image_label}(c), DSOT establishes bi-directional fine-grained correspondences between visual and textual tokens across both global and local levels. This symmetric formulation strengthens multimodal consistency and enables more precise region-text alignment. Furthermore, to incorporate spatial understanding into the segmentation process, we devise a \textbf{Composite Directional Guidance (CDG) supervision strategy} that explicitly introduces spatial constraints by constructing region-level guidance masks from textual cues. These masks provide spatial relational supervision, guiding the model to identify and localize lesions according to complex spatial expressions.

\begin{figure*}[t!]
    \centering
    \includegraphics[width=0.85\textwidth]{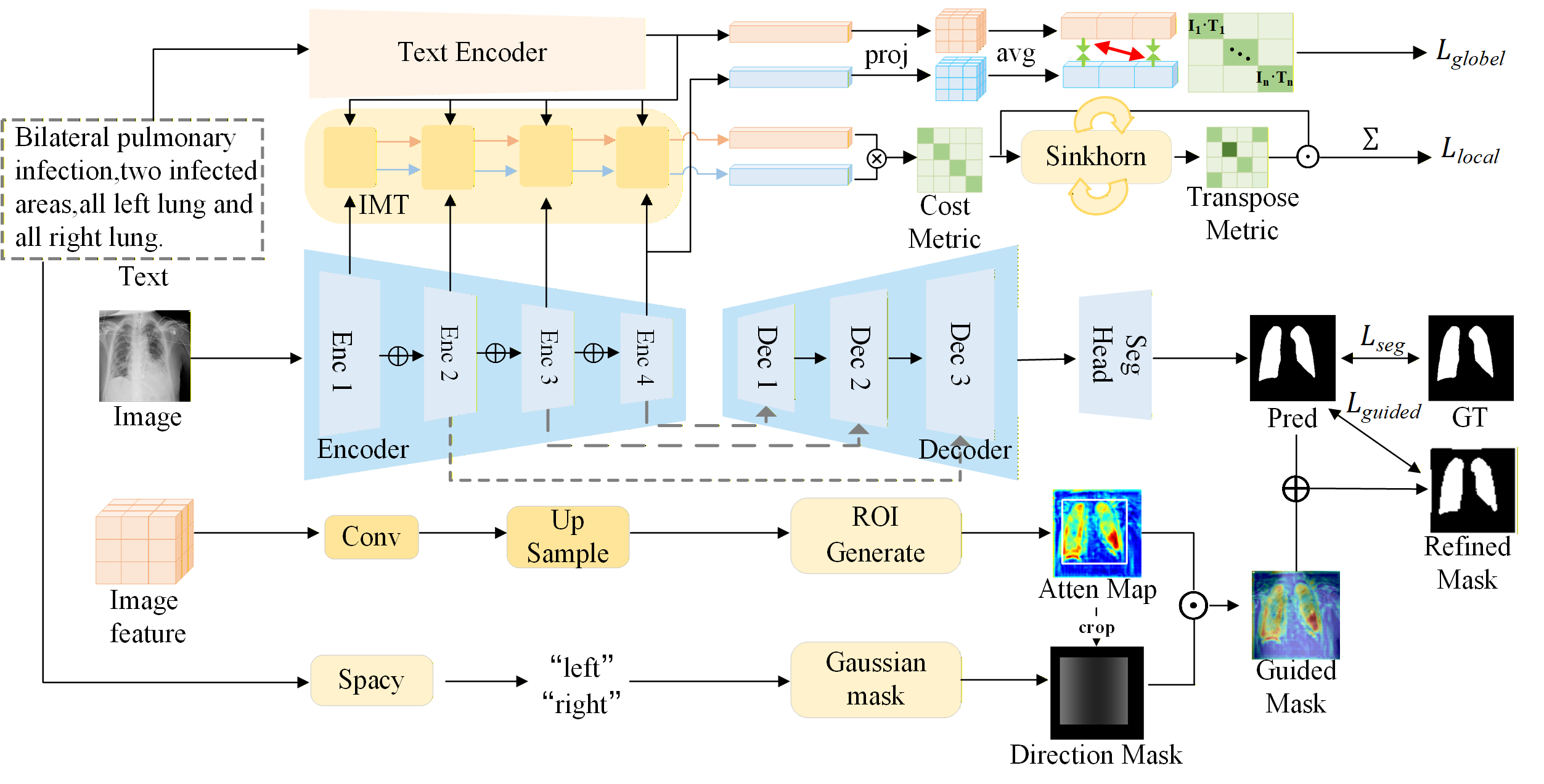}
    \caption{The overall architecture of our proposed SSA framework. Built upon a U-Net-based baseline, our approach introduces two modules. The DSOT module reinforces the alignment between visual regions and relevant textual descriptions. And the CDG module focuses on accurately segmenting lesions characterized  by spatial relational constraints.}
    \label{fig:model_architecture}
\end{figure*}

Extensive experiments on multiple public benchmarks demonstrate that the proposed framework achieves state-of-the-art performance in text-guided medical image segmentation. In particular, it consistently surpasses the baseline across diverse datasets, achieving a Dice score of 91.65 on QaTa-COV19 and a Dice score of 80.37 on MosMedData+.

The rest of the paper is organized as follows. In Section II, we detail the proposed SSA framework, especially the DSOT and CDG strategies. In Section III, qualitative and quantitative analyses are presented, along with comparisons against state-of-the-art methods. In Section IV, we summarize the proposed method.


\section{Method}
\label{sec:methodology}


We adopt a U-Net–based model~\cite{bui2024visual,zhong2023ariadne} as the baseline and construct our SSA framework upon it. As shown in Fig.~\ref{fig:model_architecture}, the framework integrates two key components to tackle the core limitations of existing methods. These components enhance the associations between image regions and multiple related textual descriptions, improving the ability of the model to identify and localize lesions according to complex clinical text. The \textbf{DSOT} module reinforces the alignment between visual regions and relevant textual descriptions. The \textbf{CDG} module focuses on accurately segmenting lesions described by spatial relational constraints.

\subsection{Dual-granularity Symmetric Optimal Transport Alignment}
\label{sec:dssot}

To resolve the ambiguity in visual-textual correspondence, our DSOT module operates at two granularities. 
At a global level, we enforce overall semantic consistency by imposing a symmetric InfoNCE contrastive loss~\cite{oord2018representation}, $\mathcal{L}_{\text{global}}$, on the averaged feature representations of the image and text tokens.
At a fine-grained level, we establish a referring, fine-grained correspondence by modeling the alignment between bi-aggregated features ($\{\mathbf{f}_{\text{img},i}\}_{i=1}^N$) and ($\{\mathbf{f}_{\text{txt},j}\}_{j=1}^L$) as an optimal transport problem. We first compute a cost matrix $\mathbf{M} \in \mathbb{R}^{N \times L}$ according to the cosine distance between token pairs:
\begin{equation}
\mathbf{M}_{ij} = 1 - \cos(\mathbf{f}_{\text{img},i}, \mathbf{f}_{\text{txt},j}).
\end{equation}

Using the Sinkhorn algorithm~\cite{jia2025adversarial}, we solve for the optimal transport problem in both the image-to-text and text-to-image directions. Furthermore, we propose the symmetric local alignment loss, $\mathcal{L}_{\text{local}}$, which minimizes the matching cost. 

These two losses are integrated into a synergistic objective, where $\mathcal{L}_{\text{global}}$ provides coarse semantic alignment guidance and $\mathcal{L}_{\text{local}}$ promotes precise fine-grained cross-modal alignment:
\begin{equation}
\mathcal{L}_{\text{align}} = \lambda_{\text{global}} \mathcal{L}_{\text{global}} + \lambda_{\text{local}} \mathcal{L}_{\text{local}},
\label{eq:align_loss}
\end{equation}
where $\lambda_{\text{global}}$ and $\lambda_{\text{local}}$ are  coefficients balancing the two loss terms.

\begin{table*}[htbp]
    \centering
    \caption{Performance Comparisons of the Proposed Method with Mono-modal and Multi-modal Methods on QaTa-COV19 and MosMedData+ Datasets. }
    \label{tab:main_results}
    \begin{tabular*}{\textwidth}{l @{\extracolsep{\fill}} lcccc}
        \toprule
        \multirow{2}{*}{\textbf{Type}} & \multirow{2}{*}{\textbf{Method}} & \multicolumn{2}{c}{\textbf{QaTa-COV19}} & \multicolumn{2}{c}{\textbf{MosMedData+}}\\
        \cmidrule(lr){3-4} \cmidrule(lr){5-6}
         & & Dice (\%) $\uparrow$ & mIoU (\%) $\uparrow$ & Dice (\%) $\uparrow$ & mIoU (\%) $\uparrow$ \\
        \midrule
        \multirow{5}{*}{Mono-Modal} 
        & U-Net \cite{ronneberger2015u}  & 83.52 & 71.71  & 67.34 & 50.76 \\
        & U-Net++ \cite{zhou2018unet++}   & 83.66 & 71.91  & 68.20 & 51.75 \\
        & AttentionUnet \cite{oktay2018attention}   & 82.13 & 69.67 & 67.93 & 51.44 \\
        & Swin UNETR \cite{hatamizadeh2021swin}   & 80.42 & 67.25 & 57.79 & 40.63 \\
        & nnU-Net \cite{isensee2021nnu}   & 84.39 & 72.35 & 69.30 & 52.53 \\
        \midrule
        \multirow{3}{*}{Multi-Modal} 
        & LViT \cite{li2023lvit}  & 83.66 & 75.11 & 74.57 & 61.33 \\
        & GuideDecoder \cite{zhong2023ariadne}   & 89.78 & 81.45 & 77.75 & 63.60 \\
        & MMI-UNet \cite{bui2024visual} & 90.88 & 83.28 & 78.42 & 64.50 \\
        & MLMoE \cite{liu2024medical}  & 91.19$\pm$0.14 & 83.81$\pm$0.09 & 77.98$\pm$0.23 & 63.90$\pm$0.17 \\
        & \textbf{Ours}  & \textbf{91.65}  & \textbf{84.60}  & \textbf{80.37} & \textbf{67.19} \\
        \bottomrule
    \end{tabular*}
\end{table*}

\begin{figure*}[htbp]
    \centering
    
    \begin{tabular}{@{}c@{\hspace{2mm}}c@{\hspace{2mm}}c@{\hspace{2mm}}c@{\hspace{2mm}}c@{\hspace{2mm}}c@{\hspace{2mm}}c@{\hspace{2mm}}c@{}}
        
        \includegraphics[width=0.105\textwidth]{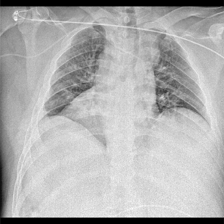} &
        \includegraphics[width=0.105\textwidth]{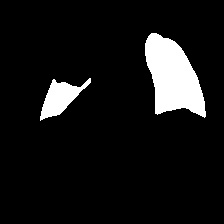} &
        \includegraphics[width=0.105\textwidth]{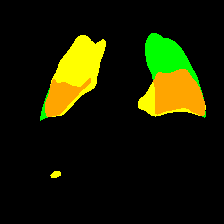} &
        \includegraphics[width=0.105\textwidth]{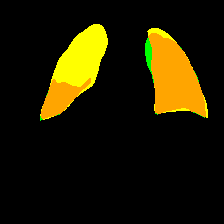} &
        \includegraphics[width=0.105\textwidth]{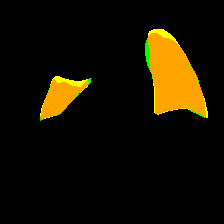} &
        \includegraphics[width=0.105\textwidth]{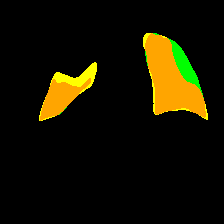} &
        \includegraphics[width=0.105\textwidth]{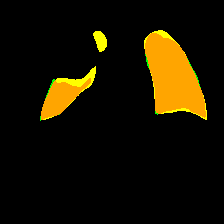} &
        \includegraphics[width=0.105\textwidth]{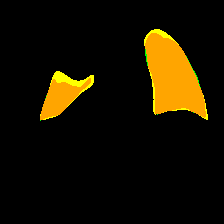} \\
        
        \includegraphics[width=0.105\textwidth]{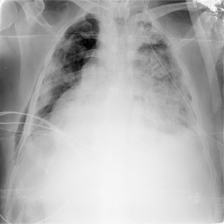} &
        \includegraphics[width=0.105\textwidth]{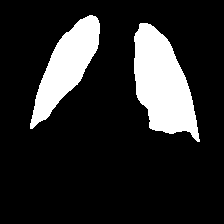} &
        \includegraphics[width=0.105\textwidth]{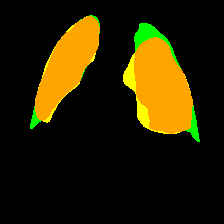} &
        \includegraphics[width=0.105\textwidth]{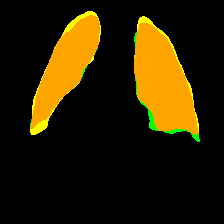} &
        \includegraphics[width=0.105\textwidth]{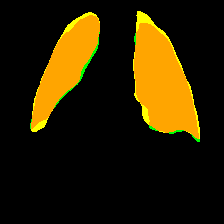} &
        \includegraphics[width=0.105\textwidth]{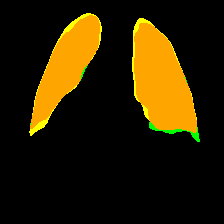} &
        \includegraphics[width=0.105\textwidth]{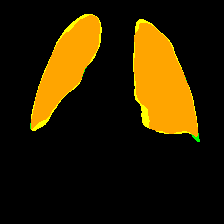} &
        \includegraphics[width=0.105\textwidth]{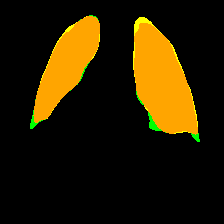} \\

        \includegraphics[width=0.105\textwidth]{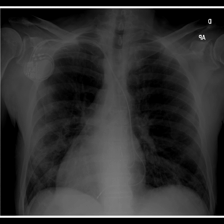} &
        \includegraphics[width=0.105\textwidth]{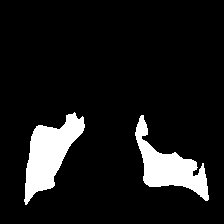} &
        \includegraphics[width=0.105\textwidth]{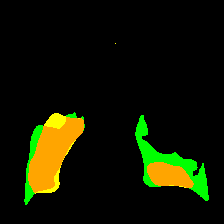} &
        \includegraphics[width=0.105\textwidth]{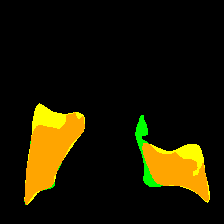} &
        \includegraphics[width=0.105\textwidth]{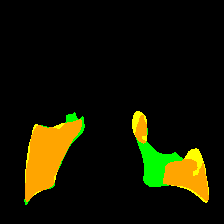} &
        \includegraphics[width=0.105\textwidth]{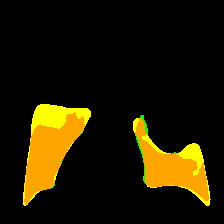} &
        \includegraphics[width=0.105\textwidth]{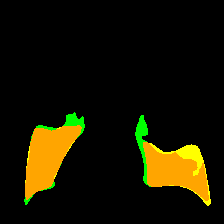} &
        \includegraphics[width=0.105\textwidth]{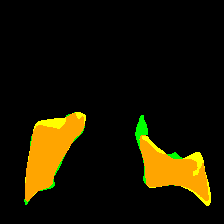} \\

        \small Raw image & \small GT & \small Swin UNETR & \small nnU-Net & \small LViT & \small GuideDecode & \small MMI-UNet  & \small Ours \\

    \end{tabular}
    
    \caption{Qualitative results on the QaTa-COV19 dataset. Our method achieves better overlap with the Ground Truth (GT) compared to other methods. Colors represent True Positive (TP), False Negative (FN), and False Positive (FP) regions in orange, green, and yellow, respectively.}
    \label{fig:qualitative_results}
\end{figure*}

\subsection{Composite Directional Guidance}
\label{sec:cdg}

To address the poor understanding of spatial expressions, the CDG module converts textual spatial cues~\cite{liu2025hybrid} into explicit supervision. This process synthesizes a guidance mask, $\mathbf{M}_{\text{guide}}$, in a dynamic, end-to-end manner. 

First, it parses directional keywords (e.g., ``lower", ``left") from the text to generate a composite 2D Gaussian mask. This prior is then localized within a region of interest delineated by a bounding box derived from the attention map, $\mathbf{A}_{\text{norm}}$. The localized prior is fused with the attention map to form the final hybrid guidance mask:
\begin{equation}
\mathbf{M}_{\text{guide}} = \mathbf{A}_{\text{norm}} \odot \mathbf{M}_{\text{pri}},
\end{equation}
where $\mathbf{M}_{\text{pri}}$ is the prior mask injected into the dynamic bounding box.

Finally, $\mathbf{M}_{\text{guide}}$ is used to regularize the initial prediction of the model, $\mathbf{P}_{\text{pred}}$, via a consistency loss. This loss compels the output of the model to be consistent with the explicit, text-derived spatial concepts:
\begin{equation}
\mathcal{L}_{\text{guide}} = \text{BCE}(\mathbf{P}_{\text{pred}}, \mathbf{P}_{\text{refined}}),
\label{eq:guidance_loss}
\end{equation}
where $\mathbf{P}_{\text{refined}}$ is a pseudo-target generated by fusing the prediction $\mathbf{P}_{\text{pred}}$ with the spatial cues $\mathbf{M}_{\text{guide}}$, and a Binary Cross-Entropy (BCE) loss is applied to constrain the prediction toward spatially consistent regions.


\section{Experiments}
\label{sec:experiments}

\subsection{Experimental Setup}
We evaluate our SSA framework on two public multi-modal medical datasets: the QaTa-COV19 and MosMedData+. Following \cite{li2023lvit}, the QaTa-COV19 dataset is divided into training, validation, and testing sets with 5716, 1429, and 2113 samples, respectively. The MosMedData+ dataset is divided into a training set with 2183 images, a validation set with 273 images and a testing set with 273 samples. The data augmentation strategies are employed during training, including random affine transformations for input images and their corresponding ground-truth masks. All models are trained using the AdamW optimizer with images resized to $224 \times 224$ pixels. Dice and mIoU are adopted to evaluate the segmentation results. Regarding the Optimal Transport hyperparameters, the entropy regularization coefficient is set to 3e-2. The Sinkhorn algorithm is configured for a maximum of 100 iterations, with a convergence threshold of 1e-2 for early stopping.

\begin{table}[htbp]
    \centering
    \caption{Ablation Study of the Proposed Components on QaTa-COV19 Dataset. }
    \label{tab:ablation_components}
    
    \begin{tabular*}{\columnwidth}{@{\extracolsep{\fill}}ccccc}
        \toprule
        \textbf{Exp. \#} & \textbf{DSOT} & \textbf{CDG} & \textbf{Dice (\%) $\uparrow$} & \textbf{mIoU (\%) $\uparrow$} \\
        \midrule
        1 & & & 90.64 & 82.89 \\
        2 &  & \checkmark& 90.92 & 83.36 \\
        3 &  \checkmark& & 91.01 & 83.50 \\
        \midrule
        4 & \checkmark & \checkmark & \textbf{91.65} & \textbf{84.60} \\
        \bottomrule
    \end{tabular*}
\end{table}

\subsection{Main Results and Analysis}
As shown in Table~\ref{tab:main_results}, our SSA framework achieves state-of-the-art (SOTA) performance on both datasets. On QaTa-COV19, SSA achieves a Dice score of \textbf{91.65\%}, significantly outperforming the previous SOTA  method (MLMoE, 91.19\%). This demonstrates the effectiveness of our DSOT and CDG. The superior performance on MosMedData+ further validates the generalization capability of our approach. The qualitative results in Fig.~\ref{fig:qualitative_results} demonstrate that our model accurately segments infected areas while suppressing irrelevant regions.

\begin{figure}[t]
    \centering
    \includegraphics[width=\columnwidth]{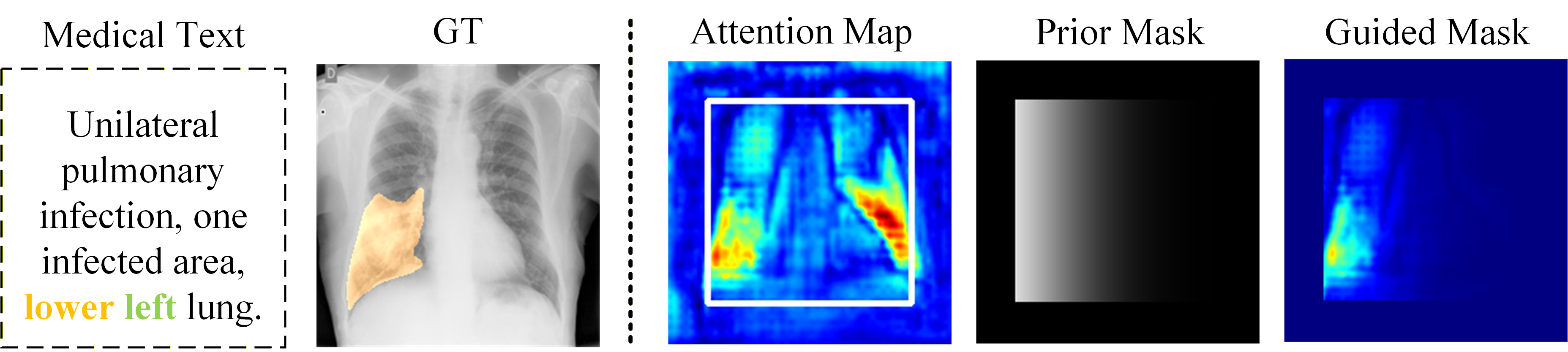}
    \caption{Visualization of the Composite Directional Guidance. The raw attention map covers regions beyond the “lower left lung", leading to potential critical deviations. Our approach generates a region-level prior mask from the text and further introduces a guided mask to enhance consistency with spatial textual descriptions.}
    \label{fig:heatmap_wide}
\end{figure}

\begin{table}[h!]
    \centering
    \caption{Impact of Training Data Size on Model Performance.}
    \label{tab:ablation_data}

    \begin{tabular*}{\columnwidth}{l @{\extracolsep{\fill}} cc}
        \toprule
        \textbf{Method} & \textbf{Dice (\%) $\uparrow$} & \textbf{mIoU (\%) $\uparrow$} \\
        \midrule
        nnUNet (100\% data) & 84.39 & 72.35 \\
        \midrule
        SSA (15\% data)     & 88.92 & 80.05 \\
        SSA (25\% data)     & 89.61 & 81.18 \\
        SSA (50\% data)     & 90.54 & 82.72 \\  
        SSA (100\% data) & \textbf{91.65} & \textbf{84.60} \\
        \bottomrule
    \end{tabular*}
\end{table}

\subsection{Ablation Study}
The ablation results on the QaTa-COV19 dataset are presented in Table~\ref{tab:ablation_components}, which validate the effectiveness of each proposed component. Compared with the baseline~\cite{bui2024visual,zhong2023ariadne}, the DSOT module improves the performance to 91.01\%, and the CDG module individually yields 90.92\%. Integrating both modules achieves the best result, with our full SSA model reaching a Dice score of \textbf{91.65\%}. As presented in
Fig.~\ref{fig:heatmap_wide}, the CDG module introduces a spatial prior mask to the raw attention map, generating an explicit guided mask to enhance consistency with complex spatial expressions. Furthermore, we assess the data efficiency by comparing SSA with a competitive mono-modal model nnU-Net, under limited-data settings in Table~\ref{tab:ablation_data}. The SSA surpasses the fully-trained nnU-Net with only 15\% of the data, demonstrating that text-guided spatial supervision substantially reduces data dependency and enhances generalization beyond purely visual models.

\section{Conclusion}
\label{sec:conclusion}
In this paper, we propose a novel text-guided medical image segmentation framework SSA. To tackle the challenge of jointly interpreting diagnostic and descriptive texts, the DSOT module introduces a symmetric optimal transport alignment mechanism, which strengthens associations between image regions and multiple relevant clinical texts. The CDG module aims to mitigate critical deviations by translating positional descriptions into explicit spatial constraints via dynamically constructing region-level guidance masks. Extensive experiments on public benchmarks demonstrate that SSA consistently surpasses state-of-the-art methods, particularly in accurately segmenting lesions described with spatial relational constraints. In future work, we are devoted to extending SSA to large multimodal models for medical image segmentation across diverse clinical scenarios.

\section*{Acknowledge}

This paper is supported by the National Natural Science Foundation of China (Grant No. 62206184).


\end{document}